\documentclass[letterpaper, 10 pt, conference]{ieeeconf}  

\IEEEoverridecommandlockouts                              

\usepackage{graphics} 
\usepackage{epsfig} 
\usepackage{mathptmx} 
\usepackage{times} 
\usepackage{amsmath} 
\usepackage{amssymb}  

\usepackage{mathtools}
\let\labelindent\relax
\usepackage{enumitem}
\usepackage{bm}

\usepackage{color}
\usepackage{xcolor}
\usepackage{booktabs}
\usepackage{multirow}
\usepackage{nicefrac}
\usepackage{url}
\usepackage{hyperref}    
\usepackage{float}

\makeatletter
\def\thanks#1{\protected@xdef\@thanks{\@thanks
        \protect\footnotetext{#1}}}
\makeatother

\title{LAC-Net: Linear-Fusion Attention-Guided Convolutional Network for Accurate Robotic Grasping  Under the Occlusion}

\author{
Jinyu Zhang$^{1*}$, Yongchong Gu$^{1*}$, Jianxiong Gao$^{1}$, Haitao Lin$^{1}$, Qiang Sun$^{2}$, \\ Xinwei Sun$^{1}$, Xiangyang Xue$^{1}$ and Yanwei Fu$^{1\dagger}$
\thanks{$*$: Equal contribution; $\dagger$: Corresponding author.}
\thanks{$^{1}$Jinyu Zhang, Yongchong Gu, Xinwei Sun and Yanwei Fu are with school of Data Science, Fudan University, China \{jyzhang23, ycgu22\}@m.fudan.edu.cn, \{sunxinwei,yanweifu\}@fudan.edu.cn;
Jianxiong Gao, Haitao Lin and Xiangyang Xue are with Fudan University, China jxgao22@m.fudan.edu.cn, \{htlin19, xyxue\}@fudan.edu.cn.
}
\thanks{$^{2}$Qiang Sun is with School of statistics and information, Shanghai University of International Business and Economics, China sunqiang@suibe.edu.cn.}
}  

\begin{document}

\maketitle
\thispagestyle{empty}
\pagestyle{empty}

\begin{abstract}
This paper addresses the challenge of perceiving complete object shapes through visual perception. While prior studies have demonstrated encouraging outcomes in segmenting the visible parts of objects within a scene, amodal segmentation, in particular, has the potential to allow robots to infer the occluded parts of objects.
To this end, this paper introduces a new framework that explores amodal segmentation for robotic grasping in cluttered scenes, thus greatly enhancing robotic grasping abilities. 
Initially, we use a conventional segmentation algorithm to detect the visible segments of the target object, which provides shape priors for completing the full object mask. 
Particularly, to explore how to utilize semantic features from RGB images and geometric information from depth images, we propose a Linear-fusion Attention-guided Convolutional Network (LAC-Net). LAC-Net utilizes the linear-fusion strategy to effectively fuse this cross-modal data, and then uses the prior visible mask as attention map to guide the network to focus on target feature locations for further complete mask recovery.
Using the amodal mask of the target object provides advantages in selecting more accurate and robust grasp points compared to relying solely on the visible segments. The results on different datasets show that our method achieves state-of-the-art performance. Furthermore, the robot experiments validate the feasibility and robustness of this method in the real world. Our code and demonstrations are available on the project page: \url{https://jrryzh.github.io/LAC-Net}. 
\end{abstract}

\section{INTRODUCTION} 

Instance segmentation is crucial for enabling robots to effectively grasp objects in predefined environments. While prior research~\cite{kirillov2023segany, xie2021unseen, xiang2021learning} has advanced visible object segment estimation, understanding occluded object parts is essential for successful grasping.

\begin{figure}[t]
\centerline{\includegraphics[width=1\columnwidth]{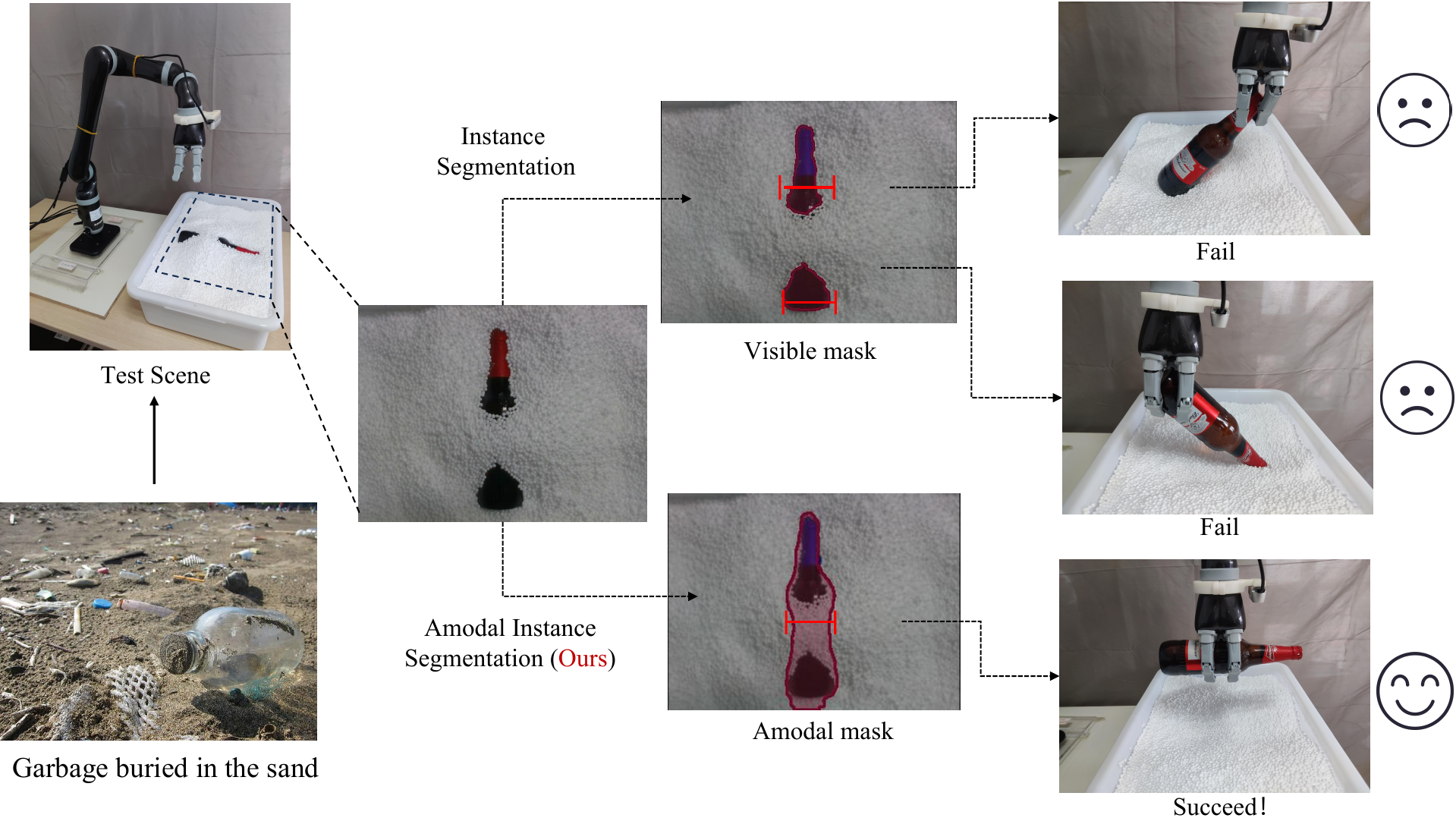}}
\vspace{-0.1in}
\caption{Comparison between the instance segmentation method and amodal segmentation for robotic grasp. We simulate the scene where the garbage is buried in the sand. Unlike conventional instance segmentation methods that predict solely the visible mask for selecting grasp points, our approach leverages the amodal mask to identify more robust grasp points for the final grasping action.  \label{fig:teaser} }
\vspace{-0.15in}
\end{figure}

For example, consider a robot cleaning debris from a beach, as illustrated in Fig.~\ref{fig:teaser}. Objects are often partially concealed by sand, making it difficult to identify their complete structure. Traditional visual masking algorithms only estimate visible parts, problematic for objects of various shapes and sizes. Relying solely on observable segments limits the information available for grasping. Therefore, acquiring comprehensive object masks enables well-informed grasp point selections, significantly enhancing grasp success rates, as shown in Fig.~\ref{fig:teaser}.

Estimating complete object masks is challenging for several reasons. Firstly, perceiving occluded areas, whether for rigid or non-rigid objects, offers many plausible yet non-unique possibilities. Secondly, real-world scenarios present complex object categories and shapes, making it difficult to pre-learn object shapes accurately. Consequently, estimating the full mask from a single-view image remains an open question.

To address the first challenge, we leverage both RGB and depth information to create robust features, reducing ambiguity through diverse modalities. For the second challenge, we utilize shape priors as initializations to guide amodal mask segmentation. Neural networks are employed to learn the potential distribution of the target object.


Formally, this paper introduces a framework integrating amodal segmentation to enhance robotic grasping in challenging, cluttered environments. Initially, we use off-the-shelf segmentation algorithms like the Segment Anything Model (SAM)~\cite{kirillov2023segany} or Mask R-CNN~\cite{he2017mask} to estimate the visible mask of the target object, providing shape priors with the RGB image. Next, our proposed LAC-Net model leverages color and depth information to transform the visible mask into a comprehensive amodal mask. Finally, we determine grasp points using RGB-aligned depth images combined with the full mask of the target object.

In summary, this work makes the following contributions:

\begin{itemize}[leftmargin=*,itemsep=0pt,topsep=0pt,parsep=0pt]
\item We present a new class-agnostic framework aimed at detecting both visible and amodal masks, enabling accurate and efficient grasping of target objects in cluttered scenes.
\item We introduce a novel component, Linear Attention Fusion, designed to fuse multi-modal information with the goal of improving the performance of amodal instance segmentation. Evaluation on the UOAIS-Sim and OSD-amodal benchmarks~\cite{back2022unseen} demonstrates state-of-the-art performance, confirming the effectiveness of our approach.
\item  Additionally, we validate the real-world applicability of amodal grasping in robotic settings, demonstrating its precision in grasping occluded target objects within complex, cluttered environments.
\end{itemize}


\section{RELATED WORKS}

\subsection{Amodal Instance Segmentation}

Amodal Instance Segmentation~\cite{zhu2017semantic} is a more complex task than standard instance segmentation~\cite{he2017mask,girshick2015fast,guo2022adaptive,li2024hfvos,guo2023openvis}, as it involves predicting not only the visible segments of objects but also their occluded shapes. Obtaining accurate masks of target objects is essential for robotic tasks such as grasping~\cite{sundermeyer2021contact,lin2022sar} and manipulating unseen objects~\cite{murali20206,lin2023pourit}. Previous research~\cite{zhang2019learning, ke2021occlusion, yang2019embodied, ling2020variational, gao2023coarse, gao2024hyper} has often focused on modeling shape priors based on shape statistics. Notably, VRSP~\cite{xiao2021amodal} introduced a shape prior module specifically for refining amodal masks. However, adapting these models to open-world scenarios, especially with complex long-tail object category distributions, remains a challenge. AISFormer~\cite{tran2022aisformer} uses transformers to capture long-range dependencies and employs multi-task training to create a comprehensive segmentation model. SaVos~\cite{yao2022self} combines spatiotemporal consistency with optical flow for amodal mask prediction, while UOAIS-ent~\cite{back2022unseen} integrates RGB-D images with multi-tasking techniques to improve amodal mask predictions for robotic grasping. Our method, using separate backbones and a linear fusion technique, effectively integrates information from both RGB and depth data, significantly enhancing the performance of amodal mask completion.

\subsection{Target-oriented Grasping in Clutter}

Robotic grasping in cluttered environments remains a formidable challenge. Significant strides have advanced the state-of-the-art in cluttered object grasping~\cite{pinto2016supersizing, mahler2017learning, mahler2017dex, kalashnikov2018scalable, wang2023wall}, particularly in generating 6DoF grasp proposals~\cite{sundermeyer2021contact,fang2020graspnet,wang2021graspness,son2022grasping}. Target-oriented grasping has also progressed, incorporating target imagery~\cite{sun2021gater}, sketches~\cite{lin2022know}, language instructions~\cite{cheang2022learning, sun2023language}, and guided demonstrations~\cite{laskey2016robot} to improve retrieval precision. Methods for target-centric grasping in cluttered contexts have been demonstrated~\cite{jang2017end,zeng2022robotic}, involving navigation through obstacles and pre-grasp operations like strategic pushing~\cite{yang2020deep,kiatos2019robust,kurenkov2020visuomotor,xu2021efficient}. Reinforcement learning has further refined these techniques~\cite{xu2023joint}. Despite advancements, the dynamic and unpredictable nature of cluttered environments—especially those with additional hindrances like snow or sand—poses unique challenges. Traditional obstacle-removal strategies often fail in these conditions. Our approach introduces a novel methodology leveraging amodal segmentation with enhanced generalization capabilities, enabling accurate identification of comprehensive object masks and facilitating the direct grasping of unseen target objects in challenging environments.


\begin{figure*}[htbp]
\centerline{\includegraphics[width=1\textwidth]{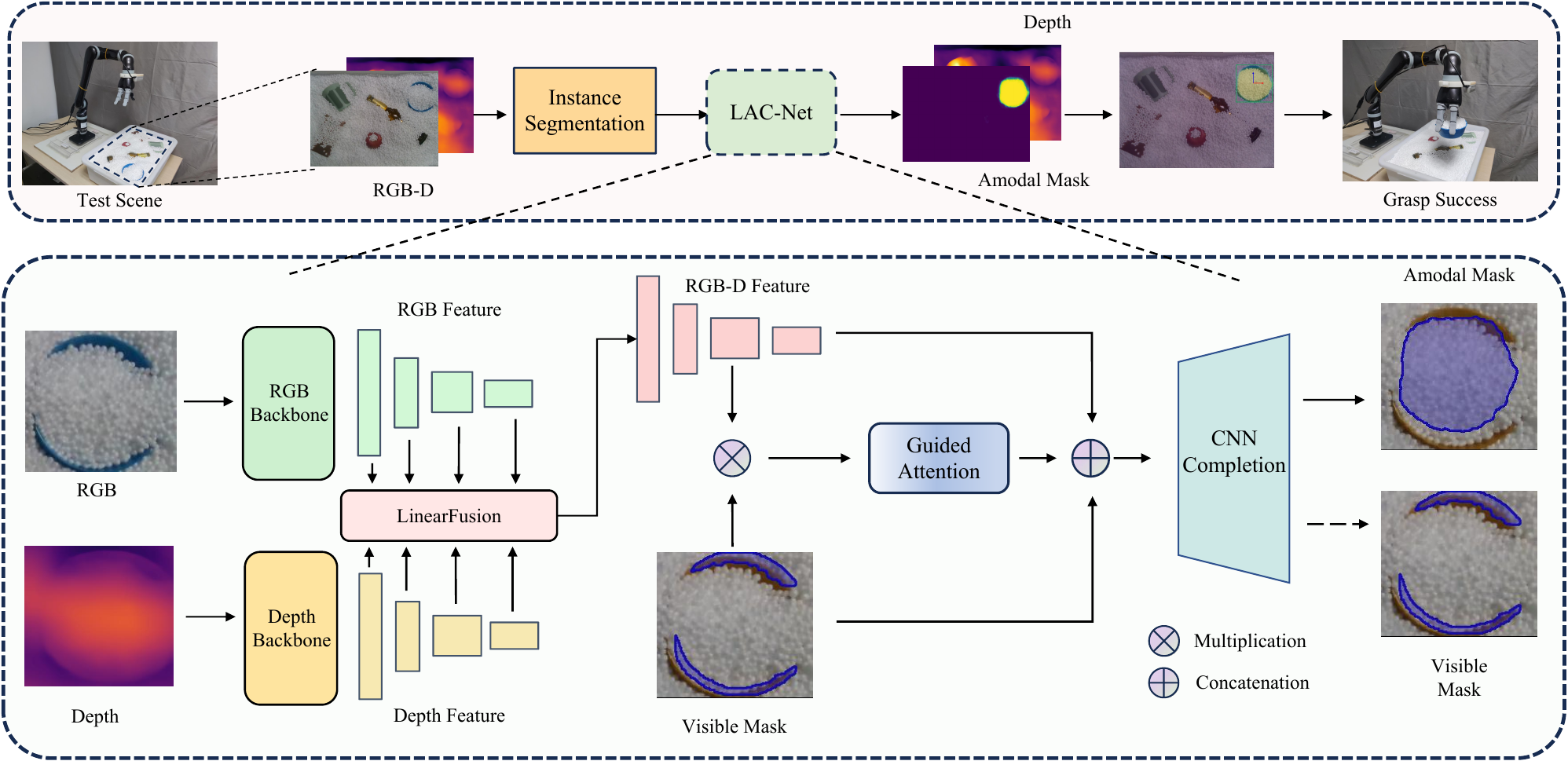}}
\vskip -0.05in
\caption{The workflow of our proposed amodal instance segmentation method for robotic grasping. Given a prompt depicting the target object within the scene, we employ Grounding DINO~\cite{liu2023grounding} in conjunction with the Segmentation Anything Model~\cite{kirillov2023segany} to localize the visible portion of the target object. Subsequently, the visible mask is fed into the coarse-to-fine amodal segmentation module to estimate the complete shape mask of the object. Finally, the robot performs a top-grasp action on the target object using the derived full mask of the target object in conjunction with depth images.\label{fig:pipeline}}
\vskip -0.2in
\end{figure*}

\section{METHOD}
\subsection{Problem Definition}
The objective of amodal instance segmentation is to simultaneously segment both the visible and occluded parts of an object. Consider an RGB-D image \( I = \{R, U\} \) capturing scene \( S \), where \( R \in \mathbb{R}^{W \times H \times 3} \) is the RGB image and \( U \in \mathbb{R}^{W \times H} \) represents the depth image. Within this scene, objects are typically comprised of two components: the visible portion and the occluded portion. We represent the mask of the visible part of an object as \( \mathbf{M}_v \) and the comprehensive amodal mask, which includes both visible and occluded sections, as \( \mathbf{M}_a \). Our framework utilizes the visible mask \( \mathbf{M}_v \), alongside the RGB-D image \( I \), to accurately infer the amodal mask \( \mathbf{M}_a \), thus achieving a complete representation of the target object.

\subsection{Framework}

Our framework consists of two main components: an instance segmentation network and a network for amodal mask completion. The instance segmentation network can be any off-the-shelf model, such as SAM~\cite{kirillov2023segany} or Mask R-CNN~\cite{he2017mask}, both of which can be trained for optimal results. High-quality instance segmentation provides better visible masks, which in turn leads to higher-quality amodal masks in our model. This relationship is further detailed in the subsequent ablation study.

\subsection{Amodal Mask Completion}
\noindent\textbf{Overview.}
The network comprises two parts: (1) RGB-D fusion backbone, (2) attention-guided completion head. First, we crop the original RGB-D input  \( I \) using the visible mask to get 256$\times$256 RGB and depth images of the object. These are processed through the RGB-D fusion backbone to generate combined RGB-D features. Next, attention maps are generated using these features and the visible mask to ignore occluders and better capture the object's details. Finally, the attention map, visible mask, and RGB-D features are merged and fed into the completion head, which outputs the completed amodal mask. This ensures an accurate representation of the object, unaffected by occlusions.

\noindent\textbf{RGB-D Fusion Backbone.}
RGB images  \( R \)  can provide color and visual texture information, while depth images can \( U \) provide spatial distance and three-dimensional structural data. Although both modalities are valuable for predicting amodal masks, efficiently leveraging them together is challenging. To tackle this, we use two separate ResNet-50 networks to independently extract features from RGB and depth inputs, retaining features from layers 1 through 4. We then fuse the corresponding features from each layer using four linear layers, resulting in a new set of features that match the original ResNet feature dimensions. Our experiments show that this approach of using dual independent backbones combined with linear fusion effectively preserves both RGB and depth features, enhancing the model's generalizability and robustness.

\noindent\textbf{Attention guided completion head}
Common amodal mask prediction networks often directly concatenate the visible mask with extracted RGB features or RGB-D features, or they perform element-wise multiplication before proceeding with further processing, such as using an upsampling CNN to obtain the amodal mask. However, we propose a more effective way to utilize the visible mask in conjunction with the features. By applying an attention mechanism, we use the visible mask as guidance to calculate attention maps in combination with the RGB-D features. These maps are then concatenated with the visible mask and RGB-D features before being fed into the subsequent completion head. This method refines the feature representation, enabling a more informed and precise completion of the amodal mask. Our completion network is primarily composed of multiple convolutional layers. It progressively interpolates and convolves the composite of previously extracted features, the mask, and the attention map. Ultimately, the network branches into two outputs: one for visible mask and another for amodal mask. The final loss function is simple, as to minimize the binary cross-entropy loss for both the visible and amodal masks simultaneously, \(\mathcal{L} := \text{BCE}(\hat{\mathbf{M}}_a, \mathbf{M}_a) + \text{BCE}(\hat{\mathbf{M}}_v, \mathbf{M}_v) \).

\subsection{Grasp Point Generation}
Given the ease of capturing RGB-D images with an RGB-D camera, we can conveniently extract the depth pixel corresponding to the target object using the estimated amodal mask. By calculating the center of this mask and utilizing the camera's intrinsic parameters, we back-project this center into a 3D point. For our grasping scenario, we employ a top-grasp strategy, concentrating solely on the object's center for robotic grasping.

\section{EXPERIMENT}
\subsection{Amodal segmentation}
\noindent\textbf{Datasets.}
We evaluate our methods on UOAIS-Sim and OSD-amodal benchmarks. UOAIS-Sim~\cite{back2022unseen} features 50,000 RGB-D images from 1,000 cluttered scenes with amodal annotations. Created using photorealistic rendering with BlenderProc, it includes 375 3D textured models of household and industrial items. Objects vary in number (1-40) and are placed on textured surfaces, captured from random camera angles. The dataset is divided into training and test sets in 9:1 and 4:1 ratios, respectively.

The OSD-amodal~\cite{richtsfeld2012segmentation} dataset consists of 111 RGB-D images capturing diverse scenes, such as Boxes, Stacked Boxes, Cylindric Objects, Mixed Objects, and Complex Scenes. It is split into 45 training and 66 validation images, with annotations including amodal masks, visible masks, and occluded masks for all objects in each scene.

\noindent\textbf{Evaluation Metric.} 
For evaluation, we firstly employ the mean Intersection over Union (mean-IoU) to assess the accuracy of the predicted amodal masks, considering both the full mask (denoted as mIoU\(_{\text{full}}\)) and the occluded region (mIoU\(_{\text{occ}}\)). The occluded mIoU is particularly insightful, offering a direct evaluation of the segmentation quality in the occluded portions of the objects, which is a critical aspect in amodal segmentation studies.

On the OSD-amodal dataset, we further measured the Overlap P/R/F, Boundary P/R/F, and \( F@\text{.75} \) for amodal and invisible masks~\cite{Xie_Yu_Mousavian_Fox_2019, Dave_Tokmakov_Ramanan_2019}. Overlap P/R/F provides insights into the total area overlap, while Boundary P/R/F reflects the precision of the boundary contours in the predicted instance masks, following Hungarian matching. The \( F@\text{.75} \) metric quantifies the percentage of segmented objects with an Overlap F-measure above 0.75. For additional insights, refer to~\cite{Xie_Yu_Mousavian_Fox_2019, Dave_Tokmakov_Ramanan_2019}. The accuracy (\( ACC_O \)) and F-measure (\( F_O \)) of occlusion classification were also calculated, where \( ACC_O = \frac{\delta}{\alpha} \), \( F_O = \frac{2P_OR_O}{P_O + R_O} \), with \( P_O = \frac{\delta}{\beta} \), and \( R_O = \frac{\delta}{\gamma} \). Here, \( \alpha \) is the total number of matched instances, \( \beta \) and \( \gamma \) represent the count of occlusion predictions and ground truth instances, respectively, and \( \delta \) signifies the correctly predicted occlusions.

\noindent\textbf{Implementation Details.}
Our amodal segmentation model is implemented using PyTorch to ensure reproducibility and facilitate comparative analysis. We adopt two segmentation strategies to conduct a fair comparison. The first strategy utilizes dataset-provided ground truth masks to gauge the model's optimal performance. The second employs the visible bounding boxes and masks detected by UOAIS-net (akin to Mask R-CNN) for a consistent baseline comparison. For preparing RGB-D images and masks, we crop them using bounding boxes of visible areas, expanded by a factor of 2, and standardize these inputs to a resolution of $256 \times 256$. We augment the masks with morphology dilation, erosion, and Gaussian blur. We use the AdamW optimizer with a learning rate of 3e-4 and batch size of 32, training for 100K iterations on the UOAIS-Sim training set. Performance is evaluated on both the UOAIS-Sim test set and the OSD-amodal dataset.

\noindent\textbf{Performance on Amodal Segmentation.}
We compared our method with state-of-the-art instance segmentation approaches, all trained on the UOAIS-Sim dataset's RGB-D data. For our method, we trained separately using depth, RGB, and RGB-D data, and benchmarked all methods on the OSD-amodal dataset. ASN and UOAIS-net directly output occlusion status, for other methods and ours, occlusion is determined based on the ratio of \( V \) to \(  A \) (occlusion \(  O = 1 \) if  \(  V/A < 0.95)\). Experimental results show that our method outperforms all baselines across all metrics. Notably, even our models trained with solely RGB or depth data achieved commendable results. When trained with RGB-D data, our method surpassed baselines in nearly all metrics, demonstrating high accuracy without a dedicated occlusion prediction branch.

\noindent\textbf{Ablation study.}
We analyzed the impact of two key factors: visible mask quality and RGB-D feature fusion methods.
\noindent\paragraph{Visible mask} We compared results using visible masks detected by UOAIS-net (similar to Mask R-CNN) against those using ground truth visible masks. As shown in Table~\ref{tab:mask_compare}, using ground truth masks significantly improved IoU on both UOAIS-Sim and OSD-amodal datasets, highlighting that higher quality visible masks enhance amodal mask completion effectiveness.

\begin{table}[ht] 
\centering
\footnotesize
\caption{Our LAC-Net V.S. UOAIS-Net by using various inputs. \{$^*$\}  demotes using Ground-truth visible mask as inputs. }
\vskip -0.1in
\setlength{\tabcolsep}{1.6mm}{
    \begin{tabular}{c|c|cc|cc}
    \toprule[1pt]
    \multirow{2}{*}{Method} & \multirow{2}{*}{In} & \multicolumn{2}{c|}{UOAIS-sim} & \multicolumn{2}{c}{OSD-amodal} \\
                            &                        & IoU full       & IoU occ       & IoU full    & IoU occ    \\ \midrule[0.5pt]
    UOAIS               & RGB-D                  & 85.51    & 53.04     & 81.73  & 45.31      \\ \midrule[0.5pt]
    LAC-Net           & RGB                      & 92.74    & 57.63    & 81.32  & 54.34            \\
    LAC-Net           & D                      & 92.41      & 50.05     & 82.10  & 48.97             \\
    LAC-Net           & RGB-D                  & 93.21      & 62.16     & 83.81  & 58.03             \\ \midrule[0.5pt]
    LAC-Net$^*$           & RGB                    & 96.57     & 73.35    & 91.41  & 70.10           \\
    LAC-Net$^*$           & D                  & 94.23      & 71.04     & 94.42  & 76.98            \\
    LAC-Net$^*$           & RGB-D              & 97.01      & 77.03     & 95.14  & 81.30             \\ \bottomrule[1pt]
    \end{tabular}
}
\vskip -0.2in
\label{tab:mask_compare}
\end{table}
    
\noindent\paragraph{RGB-D Fusion} We tested three RGB-D feature fusion methods: (1) merging RGB and depth into a 6-channel image for ResNet-50 input, (2) fusing features from ResNet-50 for RGB and depth using 1x1 convolutions, and (3) our final model's approach, fusing RGB and depth features via linear layers. Linear fusion outperformed the other methods, demonstrating robust training results and superior generalization on the unseen OSD-amodal dataset.

\begin{table}[hb]
\vskip -0.1in
\centering
\caption{Performance using different cross-modal fusion methods. }
\vskip -0.1in
\begin{tabular}{c|cc|cc}
\toprule[1pt]
                         & \multicolumn{2}{c|}{UOAIS-sim}                         & \multicolumn{2}{c}{OSD-amodal}                         \\
\multirow{-2}{*}{Method} & Iou full                              & Iou occ        & Iou full                              & Iou occ        \\ \midrule[0.5pt]
RGB-D 6 channel           & 95.71                                 & 73.72          & 94.96                                 & 80.74          \\
RGB-D 1x1conv            & 96.75                                 & 75.17          & 94.95                                 & 80.24          \\
RGB-D Linear Fusion      & {\color[HTML]{333333} \textbf{97.00}} & \textbf{77.01} & {\color[HTML]{333333} \textbf{95.05}} & \textbf{81.22} \\ \bottomrule[1pt]
\end{tabular}
\vskip -0.15in
\end{table}

\begin{table*}[htb]
\centering
\caption{Our model performances on OSD-amodal. \{\} denotes that they are predicted parallelly. \(\rightarrow\) refers the hierarchy in prediction heads. OV: Overlap \(F\), BO: Boundary \(F\)}
\vskip -0.1in
\setlength{\tabcolsep}{2.2mm}{
    \begin{tabular}{c|c|c|ccc|ccc|cc}
    \toprule[1pt]
    \multirow{2}{*}{Method} & \multirow{2}{*}{Input} & \multirow{2}{*}{Hierarchy Order}                  & \multicolumn{3}{c|}{Amodal Mask (A)}          & \multicolumn{3}{c|}{Invisible Mask (IV)}      & \multicolumn{2}{c}{Occlusion (O)} \\
                            &                        &                                                   & OV            & BO            & F@.75         & OV            & BO            & F@.75         & FO              & ACCO            \\ \midrule[0.5pt]
    Amodal MRCNN            & RGB-D                  & \{B, V, A\}                                       & 82.4          & 66.6          & 82.5          & 50.9          & 28.4          & 41.9          & 74.5            & 81.9            \\
    ORCNN                   & RGB-D                  & \{B, V, A\} $\rightarrow$ IV                      & 83.1          & 67.2          & 84.1          & 49.2          & 25.8          & 33.6          & 75.5            & 83.1            \\
    ASN                     & RGB-D                  & \{B, O\} $\rightarrow$ \{V, A\}                   & 80.7          & 67.0          & 84.5          & 42.0          & 20.7          & 38.0          & 59.1            & 65.7            \\
    UOAIS-Net               & RGB-D                  & B $\rightarrow$ V $\rightarrow$ A $\rightarrow$ O & 82.1          & 68.7          & 83.7          & 55.3          & 32.3          & 49.2          & 82.1            & 90.9            \\
    \textbf{LAC-Net(Ours)}  & RGB                    & V $\rightarrow$ A $\rightarrow$ O               & 87.6          & 73.4          & \textbf{89.7} & 55.3          & 30.3          & 38.9          & 79.7            & 85.2            \\
    \textbf{LAC-Net(Ours)}  & Depth                  & V $\rightarrow$ A $\rightarrow$ O               & 85.6          & 60.2          & 88.0          & 51.4          & 19.7          & 37.6          & 67.0            & 71.6            \\
    \textbf{LAC-Net(Ours)}  & RGB-D                  & V $\rightarrow$ A $\rightarrow$ O               & \textbf{89.0} & \textbf{79.0} & 89.4          & \textbf{64.9} & \textbf{40.6} & \textbf{60.1} & \textbf{85.0}   & \textbf{90.9}   \\ \bottomrule[1pt]
    \end{tabular}
}
\vskip -0.15in
\end{table*}

\subsection{Real-world amodal grasping}


\begin{figure}[htbp]
\centerline{\includegraphics[width=0.75\columnwidth]{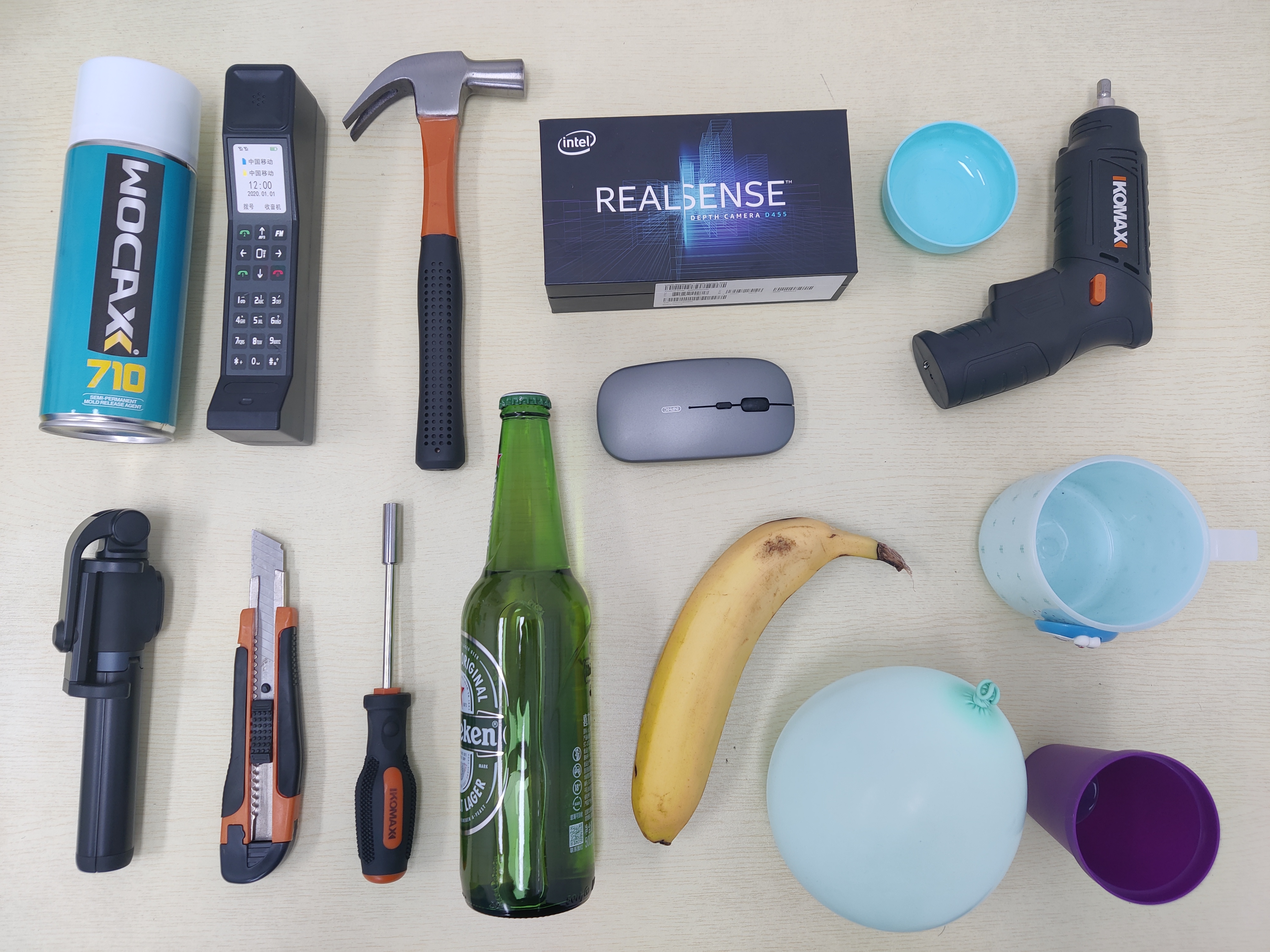}}
\caption{Test object collection used in our robotic experiment. We choose 15 instances with different shape for testing.\label{fig:grasping_targets}}
\vskip -0.1in
\end{figure}

\begin{figure*}[htbp]
\centerline{\includegraphics[width=1\textwidth]{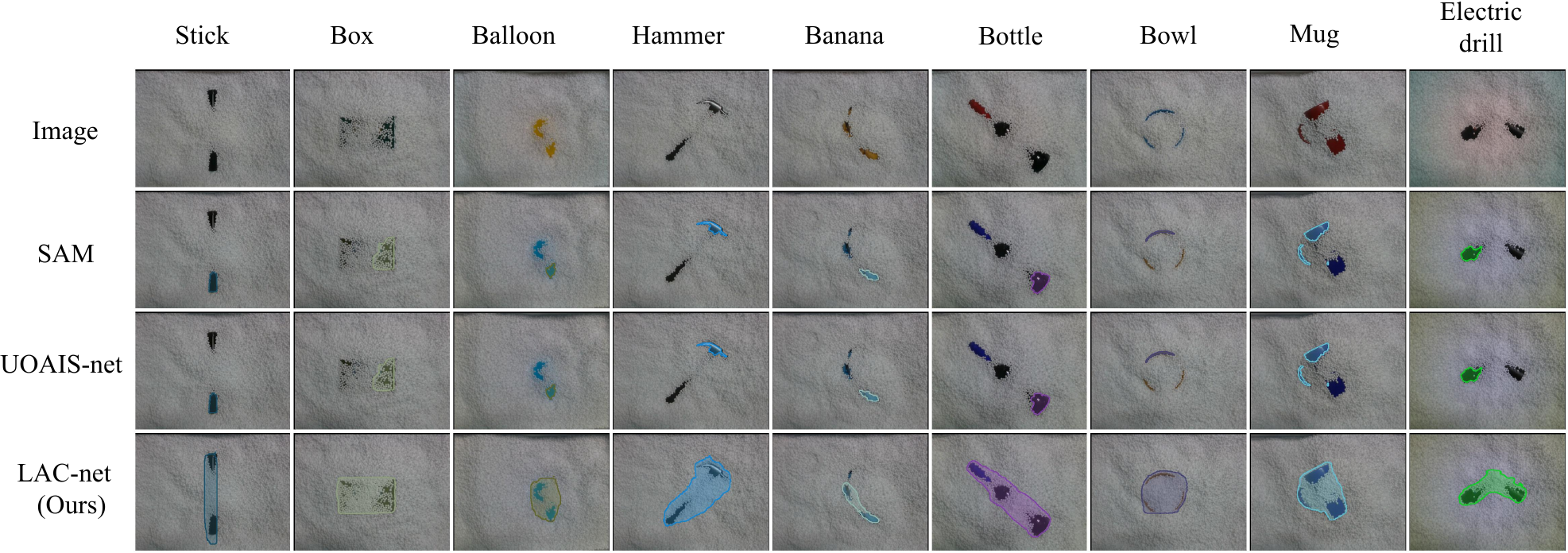}}
\vskip -0.12in
\caption{Qualitative Results of amodal segmentation in real-world Scenes: We present the original RGB image, the visible mask estimated by SAM, and the amodal mask results from both UOAIS-net and our method, respectively.\label{fig:realworld_amodal_seg}}
\vskip -0.2in
\end{figure*}

\noindent\textbf{Hardware Experimental Setup.} In our robotic experiment, we used a KINOVA Gen2 robot with a 6-DoF curved wrist and a KG-3 gripper. An arm-mounted RealSense D435 camera provides RGB-D streams of the scene. The system runs on a desktop equipped with an NVIDIA GTX2080 GPU and leverages pre-trained Grounding DINO and SAM models.

To closely simulate real-world scenarios such as snowy terrains, deserts, and marshlands, we used 5-6mm foam spheres as scene fillers to obscure the objects to be grasped. These foam spheres are highly pliable, allowing us to arrange them manually for intricate occlusion effects, thereby enhancing our ability to create complex occlusion scenarios.

\noindent\textbf{Principle of Human-robot Dialogue.} In real-world robotic experiments, experimenters engage in dialogue with the system. The system encapsulates the experimenter's preferences into target instructions in the format of ``[color] [category]'', such as ``blue bowl''. These instructions are subsequently conveyed to the visual grounding system. 

\noindent\textbf{Test Object Collections.}
To validate the effectiveness of our method, we deliberately selected fifteen diverse object types with varying shape distributions for grasp, as illustrated in Fig.~\ref{fig:grasping_targets}. These grasping targets encompass fourteen distinct object classes: stick, box, balloon, hammer, cup, banana, bottle, bowl, mug, telephone model, knife, mouse, screwdriver and electric drill. Among these, eleven classes were employed in the single-object scene, while fourteen classes were utilized in the multi-object scene grasping experiment. It is worth noting that all fourteen classes were employed for qualitative visualization purposes, enhancing the comprehensiveness of our evaluation.

\noindent\textbf{Evaluation Metrics.}
The object's full mask is divided into three equal sections along its axis. The middle section, comprising one-third of the full mask area, is designated as Region A. The remaining two sections, occupying two-thirds of the full mask area, are collectively referred to as Region B. Based on this definition, the grasping results are categorized into three types. \textit{Good}: The robot successfully grasps the object stably, with the center of the gripper positioned within Region A. \textit{Poor}: The robot successfully grasps the object stably, but the center of the gripper is located within Region B. \textit{Fail}: The robot fails to grasp the target object, either missing it entirely or being unable to maintain a grip.

\begin{figure}[tbp]
\centerline{\includegraphics[width=0.95\columnwidth]{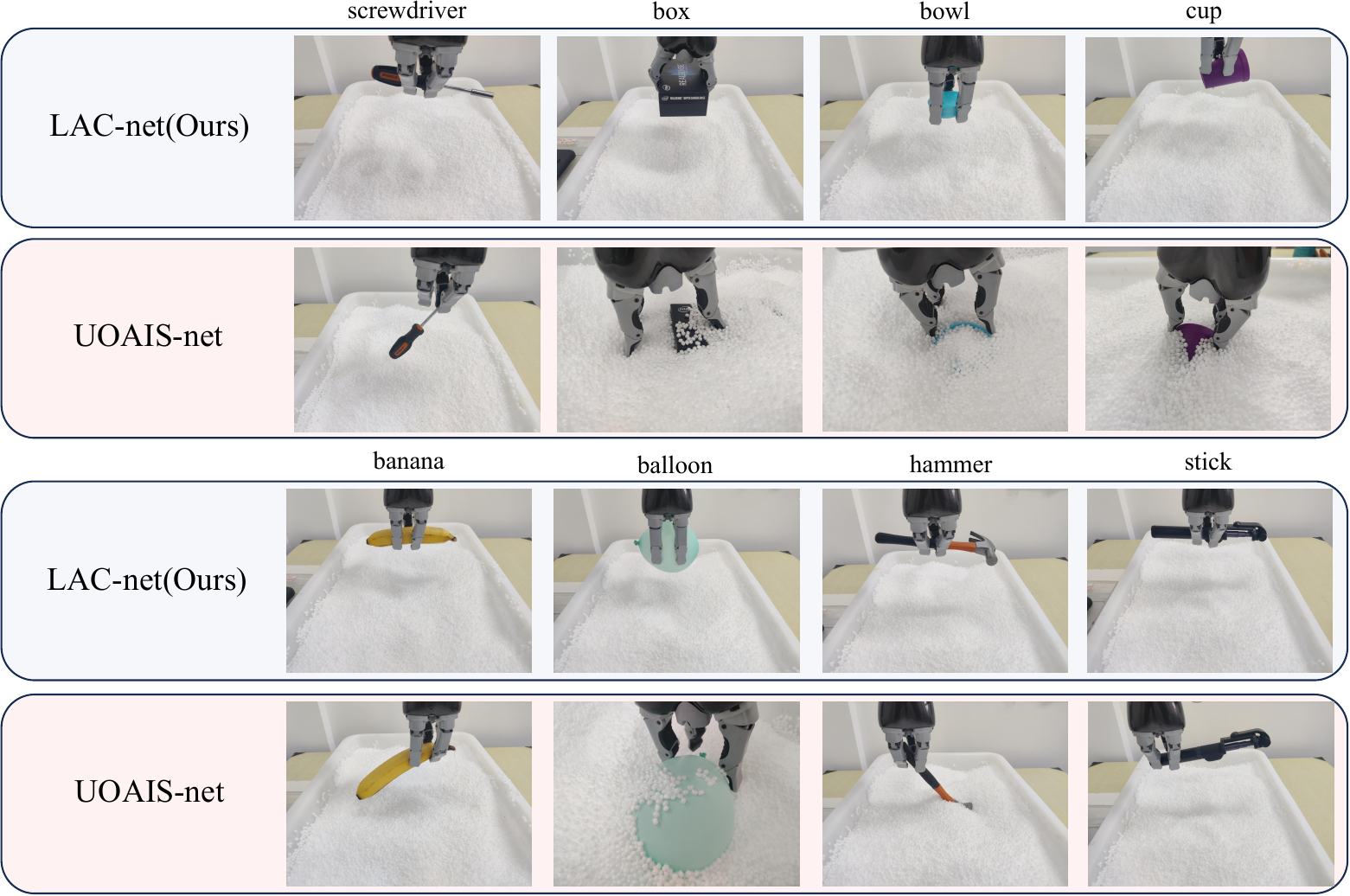}}
\vskip -0.1in
\caption{Comparison results of the robotic grasping between the baseline method and our method in \textit{single-object} scenes. In the comparison of robotic grasping methods, the baseline UOAIS-net method often tends to grasp the edge portions of the target object, primarily due to the limitations in recovering the amodal mask. In contrast, our approach excels in grasping the object's center, as our method can estimate an approximate full mask of the target object, thereby ensuring a higher rate of overall success.\label{fig:grasp_example}}
\vskip -0.25in
\end{figure}

\begin{figure}[tbp]
\centerline{\includegraphics[width=0.95\columnwidth]{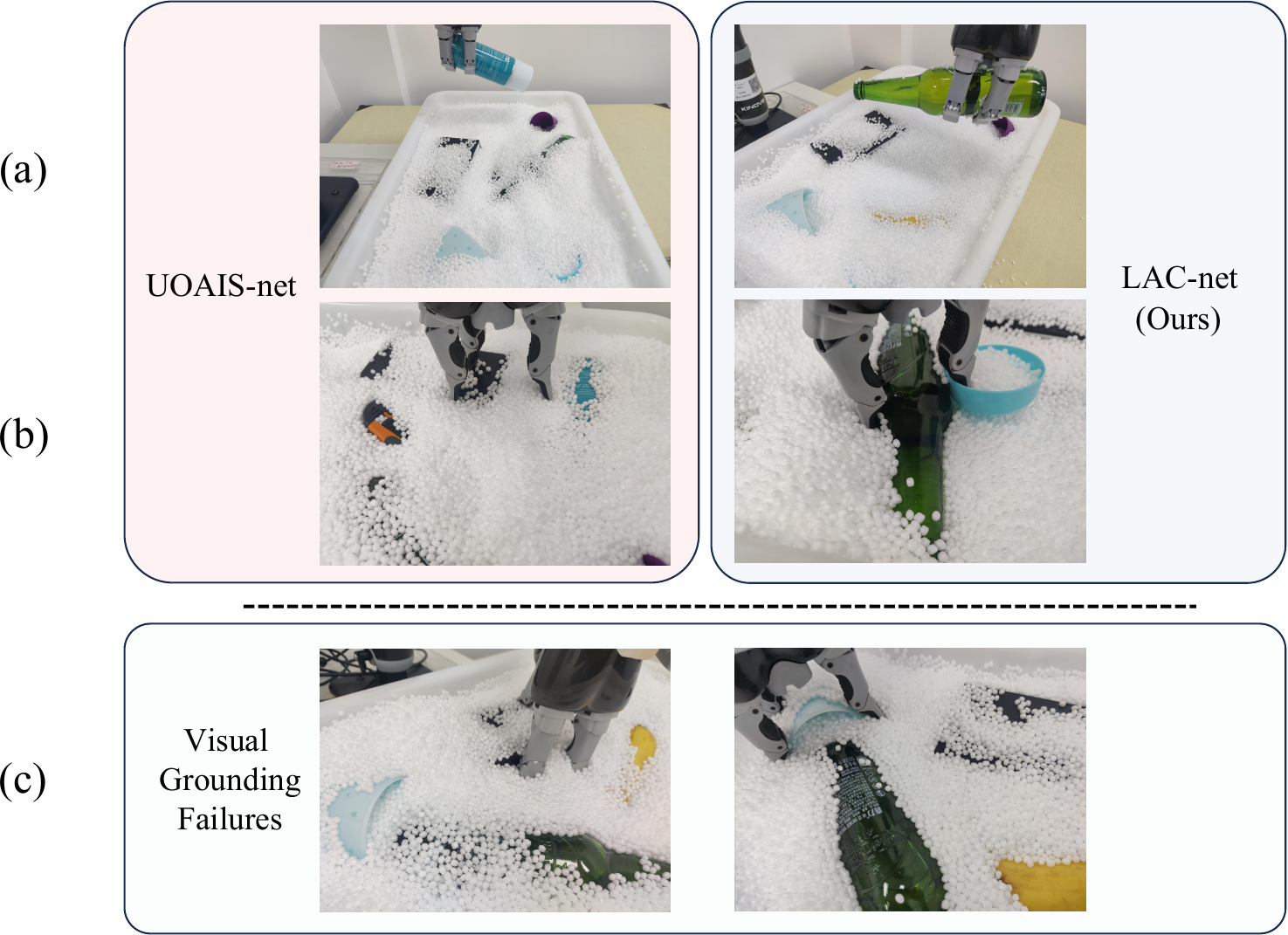}}
\vskip -0.13in
\caption{Examples of the robotic grasping between the baseline method and our method in \textit{multiple-object} scenes. (a) Successful grasps. (b) Failed grasps. (c) Instances of visual grounding failures.\label{fig:grasp_example_multi}}
\vskip -0.1in
\end{figure}

\noindent\textbf{Grasping Experiments in Single-object Scene.}
In cluttered scenes covered with foam, we conducted grasping experiments on six object categories, comparing our results with UOAIS-net under identical conditions. Detailed quantitative results in Table~\ref{tab:realword_sr} highlight our system’s performance. Our method achieved 90 successful ``Good'' grasps, significantly surpassing UOAIS-net’s 23, a margin of 67. Including both ``Good'' and ``Poor'' grasps, our method totaled 96 successful grasps, outperforming UOAIS-net’s 58 by 38 grasps.

The failures in UOAIS-net can be attributed to several factors related to object characteristics. For instance, grasping sticks and boxes often fails due to imprecision, while heavy objects like filled water bottles and hammers cause tilting and dropping when grasped at one end. Additionally, curved objects like balloons and bowls are prone to slippage, particularly near the edges, highlighting the system’s struggle with varied shapes and weights.

\begin{table}[]
\begin{centering}
\caption{Quantitative results on real-world single-object scene. G, P, F denotes good, poor, and fail grasping.\label{tab:realword_sr}}
\vskip -0.12in
\par\end{centering}
\centering{}%
\begin{tabular}{c|ccc|ccc}
\toprule[1pt]
\multirow{2}{*}{class} & \multicolumn{3}{c|}{UOAIS-net} & \multicolumn{3}{c}{Ours}         \\
                       & G       & P       & F      & G               & P     & F      \\ \midrule[0.5pt]
Stick                   & 6/20    & 14/20   & 0/20   & \textbf{17/20}  & 3/20  & 0/0    \\
Box                    & 8/20    & 11/20   & 1/20   & \textbf{20/20}  & 0/20  & 0/20   \\
Bottle                 & 3/20    & 7/20    & 10/20  & \textbf{18/20}  & 1/20  & 1/20   \\
Hammer                 & 1/20    & 3/20    & 16/20  & \textbf{11/20}  & 2/20  & 7/20   \\
Balloon                & 2/20    & 0/20    & 18/20  & \textbf{8/20}   & 0/20  & 12/20  \\
Bowl                   & 3/20    & 0/20    & 17/20  & \textbf{16/20}  & 0/20  & 4/20   \\
\textbf{All}           & 23/120  & 35/120  & 62/120 & \textbf{90/120} & 6/120 & 24/120 \\ \bottomrule[1pt]
\end{tabular}
\vskip -0.25in
\end{table}

\noindent\textbf{Grasping Experiment in Multi-object Scene}
Multi-object scenes present greater challenges than single-object scenes due to potential detection interference and the risk of collisions between the gripper and nearby objects. Table~\ref{tab:realword_sr_multi} shows our method achieving 58 successful ``good'' grasps compared to UOAIS-net's 18, a margin of 40. Overall, we achieve 63 total successful grasps, surpassing UOAIS-net's 43 by 20.

Fig.~\ref{fig:grasp_example_multi} illustrates varied outcomes of our method. In (a), we show examples of successful object grasping by UOAIS-net and LAC-net. In (b), we explore cases where close-proximity objects obstruct the robot's gripper, leading to failure, despite our method's accurate amodal mask generation. Our method significantly outperforms UOAIS-net by generating central grasp points, reducing peripheral grasp failures.

In (c), we showed some visual grounding errors, such as misidentifying ``yellow bottle'' when the system focuses on a red mug, highlighting challenges in complex environments. Overall, our method advances robotic grasping by accurately identifying central grasp points, despite issues in object proximity and visual grounding.

\begin{table}[]
\begin{centering}
\caption{Quantitative results on real-world multiple-object scene. G, P, F denotes good, poor and fail grasping.\label{tab:realword_sr_multi}}
\vskip -0.12in
\par\end{centering}
\centering{}%
\begin{tabular}{c|ccc|ccc}
\toprule[1pt]
\multirow{2}{*}{class} & \multicolumn{3}{c|}{UOAIS-net} & \multicolumn{3}{c}{Ours}         \\
                       & G       & P       & F      & G               & P     & F      \\ \midrule[0.5pt]
Stick                   & 4/20    & 7/20    & 9/20   & \textbf{9/20}   & 1/20  & 10/20  \\
Box                    & 3/20    & 7/20    & 10/20  & \textbf{15/20}  & 0/20  & 5/20   \\
Bottle                 & 4/20    & 3/20    & 13/20  & \textbf{14/20}  & 0/20  & 6/20   \\
Mug                    & 5/20    & 8/20    & 7/20   & \textbf{13/20}  & 4/20  & 3/20   \\
Bowl                   & 2/20    & 0/20    & 18/20  & \textbf{7/20}   & 0/20  & 13/20  \\
\textbf{All}           & 18/120  & 25/120  & 57/120 & \textbf{58/120} & 5/120 & 37/120 \\ \bottomrule[1pt]
\end{tabular}
\vskip -0.2in
\end{table}

\noindent\textbf{Qualitative Results}
As illustrated in Fig.~\ref{fig:realworld_amodal_seg}, our amodal segmentation technique consistently yields higher quality masks compared to those produced by UOAIS-net. We find that UOAIS-net often mistakes the visible mask for the entire object, leading to a propensity for grasping at the extremities of objects. In contrast, our method generates more comprehensive and precise amodal masks. These improved masks better guide the robotic arm to grasp at the object's center, resulting in a higher success rate and improved grasping quality.

In Fig.~\ref{fig:grasp_example}, we demonstrate the efficacy of our system with examples of single-object grasping in cluttered settings. These examples highlight that successful grasping in such scenarios hinges on accurately targeting the obscured centers of objects, rather than just their visible peripheries.

\section{CONCLUSIONS}
In this work, we introduce a novel framework for amodal instance segmentation. By extracting and linearly fusing RGB and depth features, we obtain robust RGB-D characteristics. Utilizing the visible mask, we calculate guided attention, which, through a convolutional completion network, yields the completed amodal mask. Our experiments demonstrate that our method achieves state-of-the-art (SOTA) results on UOAIS-Sim and OSD-amodal benchmarks. Furthermore, we have applied our method to a physical Kinova robot for conducting robotic grasping tasks on objects submerged in white foam. The experimental outcomes underscore our method's exceptional real-world amodal grasping capabilities, showcasing its particular effectiveness in applications like debris removal by cleaning robots in sandy contexts. Future work aims to delve further into exploiting our method's potential for beach cleaning scenarios.

\section*{Acknowledgements}
The computations in this research were performed using the CFFF platform of Fudan University.



\bibliographystyle{./IEEEtran.bst}
\bibliography{./IEEEexample.bib}

\begin{thebibliography}{10}
\providecommand{\url}[1]{#1}
\csname url@rmstyle\endcsname
\providecommand{\newblock}{\relax}
\providecommand{\bibinfo}[2]{#2}
\providecommand\BIBentrySTDinterwordspacing{\spaceskip=0pt\relax}
\providecommand\BIBentryALTinterwordstretchfactor{4}
\providecommand\BIBentryALTinterwordspacing{\spaceskip=\fontdimen2\font plus
\BIBentryALTinterwordstretchfactor\fontdimen3\font minus \fontdimen4\font\relax}
\providecommand\BIBforeignlanguage[2]{{%
\expandafter\ifx\csname l@#1\endcsname\relax
\typeout{** WARNING: IEEEtran.bst: No hyphenation pattern has been}%
\typeout{** loaded for the language `#1'. Using the pattern for}%
\typeout{** the default language instead.}%
\else
\language=\csname l@#1\endcsname
\fi
#2}}

\bibitem{kirillov2023segany}
A.~Kirillov, E.~Mintun, N.~Ravi, H.~Mao, C.~Rolland, L.~Gustafson, T.~Xiao, S.~Whitehead, A.~C. Berg, W.-Y. Lo, P.~Doll{\'a}r, and R.~Girshick, ``Segment anything,'' \emph{arXiv:2304.02643}, 2023.

\bibitem{xie2021unseen}
C.~Xie, Y.~Xiang, A.~Mousavian, and D.~Fox, ``Unseen object instance segmentation for robotic environments,'' \emph{IEEE Transactions on Robotics}, vol.~37, no.~5, pp. 1343--1359, 2021.

\bibitem{xiang2021learning}
Y.~Xiang, C.~Xie, A.~Mousavian, and D.~Fox, ``Learning rgb-d feature embeddings for unseen object instance segmentation,'' in \emph{Conference on Robot Learning}.\hskip 1em plus 0.5em minus 0.4em\relax PMLR, 2021, pp. 461--470.

\bibitem{he2017mask}
\BIBentryALTinterwordspacing
K.~He, G.~Gkioxari, P.~Doll{\'{a}}r, and R.~B. Girshick, ``Mask {R-CNN},'' \emph{CoRR}, vol. abs/1703.06870, 2017. [Online]. Available: \url{http://arxiv.org/abs/1703.06870}
\BIBentrySTDinterwordspacing

\bibitem{back2022unseen}
S.~Back, J.~Lee, T.~Kim, S.~Noh, R.~Kang, S.~Bak, and K.~Lee, ``Unseen object amodal instance segmentation via hierarchical occlusion modeling,'' in \emph{2022 International Conference on Robotics and Automation (ICRA)}.\hskip 1em plus 0.5em minus 0.4em\relax IEEE, 2022, pp. 5085--5092.

\bibitem{zhu2017semantic}
Y.~Zhu, Y.~Tian, D.~Metaxas, and P.~Doll{\'a}r, ``Semantic amodal segmentation,'' in \emph{Proceedings of the IEEE conference on computer vision and pattern recognition}, 2017, pp. 1464--1472.

\bibitem{girshick2015fast}
R.~Girshick, ``Fast r-cnn,'' in \emph{Proceedings of the IEEE international conference on computer vision}, 2015, pp. 1440--1448.

\bibitem{guo2022adaptive}
P.~Guo, W.~Zhang, X.~Li, and W.~Zhang, ``Adaptive online mutual learning bi-decoders for video object segmentation,'' \emph{IEEE Transactions on Image Processing}, vol.~31, pp. 7063--7077, 2022.

\bibitem{li2024hfvos}
W.~Li, J.~Fan, P.~Guo, L.~Hong, and W.~Zhang, ``Hfvos: History-future integrated dynamic memory for video object segmentation,'' \emph{IEEE Transactions on Circuits and Systems for Video Technology}, 2024.

\bibitem{guo2023openvis}
P.~Guo, T.~Huang, P.~He, X.~Liu, T.~Xiao, Z.~Chen, and W.~Zhang, ``Openvis: Open-vocabulary video instance segmentation,'' \emph{arXiv preprint arXiv:2305.16835}, 2023.

\bibitem{sundermeyer2021contact}
M.~Sundermeyer, A.~Mousavian, R.~Triebel, and D.~Fox, ``Contact-graspnet: Efficient 6-dof grasp generation in cluttered scenes,'' in \emph{2021 IEEE International Conference on Robotics and Automation (ICRA)}.\hskip 1em plus 0.5em minus 0.4em\relax IEEE, 2021, pp. 13\,438--13\,444.

\bibitem{lin2022sar}
H.~Lin, Z.~Liu, C.~Cheang, Y.~Fu, G.~Guo, and X.~Xue, ``Sar-net: Shape alignment and recovery network for category-level 6d object pose and size estimation,'' in \emph{Proceedings of the IEEE/CVF conference on computer vision and pattern recognition}, 2022, pp. 6707--6717.

\bibitem{murali20206}
A.~Murali, A.~Mousavian, C.~Eppner, C.~Paxton, and D.~Fox, ``6-dof grasping for target-driven object manipulation in clutter,'' in \emph{2020 IEEE International Conference on Robotics and Automation (ICRA)}.\hskip 1em plus 0.5em minus 0.4em\relax IEEE, 2020, pp. 6232--6238.

\bibitem{lin2023pourit}
H.~Lin, Y.~Fu, and X.~Xue, ``Pourit!: Weakly-supervised liquid perception from a single image for visual closed-loop robotic pouring,'' in \emph{Proceedings of the IEEE/CVF International Conference on Computer Vision}, 2023, pp. 241--251.

\bibitem{zhang2019learning}
Z.~Zhang, A.~Chen, L.~Xie, J.~Yu, and S.~Gao, ``Learning semantics-aware distance map with semantics layering network for amodal instance segmentation,'' in \emph{Proceedings of the 27th ACM International Conference on Multimedia}, 2019, pp. 2124--2132.

\bibitem{ke2021occlusion}
L.~Ke, Y.-W. Tai, and C.-K. Tang, ``Occlusion-aware video object inpainting,'' in \emph{Proceedings of the IEEE/CVF International Conference on Computer Vision}, 2021, pp. 14\,468--14\,478.

\bibitem{yang2019embodied}
J.~Yang, Z.~Ren, M.~Xu, X.~Chen, D.~J. Crandall, D.~Parikh, and D.~Batra, ``Embodied amodal recognition: Learning to move to perceive objects,'' in \emph{Proceedings of the IEEE/CVF International Conference on Computer Vision}, 2019, pp. 2040--2050.

\bibitem{ling2020variational}
H.~Ling, D.~Acuna, K.~Kreis, S.~W. Kim, and S.~Fidler, ``Variational amodal object completion,'' \emph{Advances in Neural Information Processing Systems}, vol.~33, pp. 16\,246--16\,257, 2020.

\bibitem{gao2023coarse}
J.~Gao, X.~Qian, Y.~Wang, T.~Xiao, T.~He, Z.~Zhang, and Y.~Fu, ``Coarse-to-fine amodal segmentation with shape prior,'' in \emph{Proceedings of the IEEE/CVF International Conference on Computer Vision}, 2023, pp. 1262--1271.

\bibitem{gao2024hyper}
J.~Gao, X.~Qian, L.~Liang, J.~Han, and Y.~Fu, ``Hyper-transformer for amodal completion,'' \emph{arXiv preprint arXiv:2405.19949}, 2024.

\bibitem{xiao2021amodal}
Y.~Xiao, Y.~Xu, Z.~Zhong, W.~Luo, J.~Li, and S.~Gao, ``Amodal segmentation based on visible region segmentation and shape prior,'' in \emph{Proceedings of the AAAI Conference on Artificial Intelligence}, vol.~35, no.~4, 2021, pp. 2995--3003.

\bibitem{tran2022aisformer}
M.~Tran, K.~Vo, K.~Yamazaki, A.~Fernandes, M.~Kidd, and N.~Le, ``Aisformer: Amodal instance segmentation with transformer,'' \emph{arXiv preprint arXiv:2210.06323}, 2022.

\bibitem{yao2022self}
J.~Yao, Y.~Hong, C.~Wang, T.~Xiao, T.~He, F.~Locatello, D.~Wipf, Y.~Fu, and Z.~Zhang, ``Self-supervised amodal video object segmentation,'' \emph{arXiv preprint arXiv:2210.12733}, 2022.

\bibitem{pinto2016supersizing}
L.~Pinto and A.~Gupta, ``Supersizing self-supervision: Learning to grasp from 50k tries and 700 robot hours,'' in \emph{2016 IEEE international conference on robotics and automation (ICRA)}.\hskip 1em plus 0.5em minus 0.4em\relax IEEE, 2016, pp. 3406--3413.

\bibitem{mahler2017learning}
J.~Mahler and K.~Goldberg, ``Learning deep policies for robot bin picking by simulating robust grasping sequences,'' in \emph{Conference on robot learning}.\hskip 1em plus 0.5em minus 0.4em\relax PMLR, 2017, pp. 515--524.

\bibitem{mahler2017dex}
J.~Mahler, J.~Liang, S.~Niyaz, M.~Laskey, R.~Doan, X.~Liu, J.~A. Ojea, and K.~Goldberg, ``Dex-net 2.0: Deep learning to plan robust grasps with synthetic point clouds and analytic grasp metrics,'' \emph{arXiv preprint arXiv:1703.09312}, 2017.

\bibitem{kalashnikov2018scalable}
D.~Kalashnikov, A.~Irpan, P.~Pastor, J.~Ibarz, A.~Herzog, E.~Jang, D.~Quillen, E.~Holly, M.~Kalakrishnan, V.~Vanhoucke, \emph{et~al.}, ``Scalable deep reinforcement learning for vision-based robotic manipulation,'' in \emph{Conference on Robot Learning}.\hskip 1em plus 0.5em minus 0.4em\relax PMLR, 2018, pp. 651--673.

\bibitem{wang2023wall}
T.~Wang, Y.~Li, H.~Lin, X.~Xue, and Y.~Fu, ``Wall-e: Embodied robotic waiter load lifting with large language model,'' \emph{arXiv preprint arXiv:2308.15962}, 2023.

\bibitem{fang2020graspnet}
H.-S. Fang, C.~Wang, M.~Gou, and C.~Lu, ``Graspnet-1billion: A large-scale benchmark for general object grasping,'' in \emph{Proceedings of the IEEE/CVF conference on computer vision and pattern recognition}, 2020, pp. 11\,444--11\,453.

\bibitem{wang2021graspness}
C.~Wang, H.-S. Fang, M.~Gou, H.~Fang, J.~Gao, and C.~Lu, ``Graspness discovery in clutters for fast and accurate grasp detection,'' in \emph{Proceedings of the IEEE/CVF International Conference on Computer Vision}, 2021, pp. 15\,964--15\,973.

\bibitem{son2022grasping}
D.~Son, ``Grasping as inference: Reactive grasping in heavily cluttered environment,'' \emph{IEEE Robotics and Automation Letters}, vol.~7, no.~3, pp. 7193--7200, 2022.

\bibitem{sun2021gater}
M.~Sun and Y.~Gao, ``Gater: Learning grasp-action-target embeddings and relations for task-specific grasping,'' \emph{IEEE Robotics and Automation Letters}, vol.~7, no.~1, pp. 618--625, 2021.

\bibitem{lin2022know}
H.~Lin, C.~Cheang, Y.~Fu, and X.~Xue, ``I know what you draw: Learning grasp detection conditioned on a few freehand sketches,'' in \emph{2022 International Conference on Robotics and Automation (ICRA)}.\hskip 1em plus 0.5em minus 0.4em\relax IEEE, 2022, pp. 8417--8423.

\bibitem{cheang2022learning}
C.~Cheang, H.~Lin, Y.~Fu, and X.~Xue, ``Learning 6-dof object poses to grasp category-level objects by language instructions,'' in \emph{2022 International Conference on Robotics and Automation (ICRA)}.\hskip 1em plus 0.5em minus 0.4em\relax IEEE, 2022, pp. 8476--8482.

\bibitem{sun2023language}
Q.~Sun, H.~Lin, Y.~Fu, Y.~Fu, and X.~Xue, ``Language guided robotic grasping with fine-grained instructions,'' in \emph{2023 IEEE/RSJ International Conference on Intelligent Robots and Systems (IROS)}.\hskip 1em plus 0.5em minus 0.4em\relax IEEE, 2023, pp. 1319--1326.

\bibitem{laskey2016robot}
M.~Laskey, J.~Lee, C.~Chuck, D.~Gealy, W.~Hsieh, F.~T. Pokorny, A.~D. Dragan, and K.~Goldberg, ``Robot grasping in clutter: Using a hierarchy of supervisors for learning from demonstrations,'' in \emph{2016 IEEE international conference on automation science and engineering (CASE)}.\hskip 1em plus 0.5em minus 0.4em\relax IEEE, 2016, pp. 827--834.

\bibitem{jang2017end}
E.~Jang, S.~Vijayanarasimhan, P.~Pastor, J.~Ibarz, and S.~Levine, ``End-to-end learning of semantic grasping,'' \emph{arXiv preprint arXiv:1707.01932}, 2017.

\bibitem{zeng2022robotic}
A.~Zeng, S.~Song, K.-T. Yu, E.~Donlon, F.~R. Hogan, M.~Bauza, D.~Ma, O.~Taylor, M.~Liu, E.~Romo, \emph{et~al.}, ``Robotic pick-and-place of novel objects in clutter with multi-affordance grasping and cross-domain image matching,'' \emph{The International Journal of Robotics Research}, vol.~41, no.~7, pp. 690--705, 2022.

\bibitem{yang2020deep}
Y.~Yang, H.~Liang, and C.~Choi, ``A deep learning approach to grasping the invisible,'' \emph{IEEE Robotics and Automation Letters}, vol.~5, no.~2, pp. 2232--2239, 2020.

\bibitem{kiatos2019robust}
M.~Kiatos and S.~Malassiotis, ``Robust object grasping in clutter via singulation,'' in \emph{2019 International Conference on Robotics and Automation (ICRA)}.\hskip 1em plus 0.5em minus 0.4em\relax IEEE, 2019, pp. 1596--1600.

\bibitem{kurenkov2020visuomotor}
A.~Kurenkov, J.~Taglic, R.~Kulkarni, M.~Dominguez-Kuhne, A.~Garg, R.~Mart{\'\i}n-Mart{\'\i}n, and S.~Savarese, ``Visuomotor mechanical search: Learning to retrieve target objects in clutter,'' in \emph{2020 IEEE/RSJ International Conference on Intelligent Robots and Systems (IROS)}.\hskip 1em plus 0.5em minus 0.4em\relax IEEE, 2020, pp. 8408--8414.

\bibitem{xu2021efficient}
K.~Xu, H.~Yu, Q.~Lai, Y.~Wang, and R.~Xiong, ``Efficient learning of goal-oriented push-grasping synergy in clutter,'' \emph{IEEE Robotics and Automation Letters}, vol.~6, no.~4, pp. 6337--6344, 2021.

\bibitem{xu2023joint}
K.~Xu, S.~Zhao, Z.~Zhou, Z.~Li, H.~Pi, Y.~Zhu, Y.~Wang, and R.~Xiong, ``A joint modeling of vision-language-action for target-oriented grasping in clutter,'' \emph{arXiv preprint arXiv:2302.12610}, 2023.

\bibitem{liu2023grounding}
S.~Liu, Z.~Zeng, T.~Ren, F.~Li, H.~Zhang, J.~Yang, C.~Li, J.~Yang, H.~Su, J.~Zhu, \emph{et~al.}, ``Grounding dino: Marrying dino with grounded pre-training for open-set object detection,'' \emph{arXiv preprint arXiv:2303.05499}, 2023.

\bibitem{richtsfeld2012segmentation}
A.~Richtsfeld, T.~M{\"o}rwald, J.~Prankl, M.~Zillich, and M.~Vincze, ``Segmentation of unknown objects in indoor environments,'' in \emph{2012 IEEE/RSJ International Conference on Intelligent Robots and Systems}.\hskip 1em plus 0.5em minus 0.4em\relax IEEE, 2012, pp. 4791--4796.

\bibitem{Xie_Yu_Mousavian_Fox_2019}
C.~Xie, X.~Yu, A.~Mousavian, and D.~Fox, ``\BIBforeignlanguage{en-US}{The best of both modes: Separately leveraging rgb and depth for unseen object instance segmentation},'' \emph{\BIBforeignlanguage{en-US}{Cornell University - arXiv,Cornell University - arXiv}}, Jul 2019.

\bibitem{Dave_Tokmakov_Ramanan_2019}
A.~Dave, P.~Tokmakov, and D.~Ramanan, ``\BIBforeignlanguage{en-US}{Towards segmenting anything that moves},'' \emph{\BIBforeignlanguage{en-US}{arXiv: Computer Vision and Pattern Recognition,arXiv: Computer Vision and Pattern Recognition}}, Feb 2019.

\end{thebibliography}

\end{document}